# A convolutional neural network reaches optimal sensitivity for detecting some, but not all, patterns

Fabian H. Reith
Brian A. Wandell
Psychology Department, Stanford University, Stanford, CA 94305, USA

**Contact information:**
Fabian.H.Reith@gmail.com
wandell@stanford.edu

**Acknowledgements:** We thank David Donoho, Zhenyi Liu, Zheng Lyu, and Daniel Yamins for useful discussions.

## Abstract

We investigate the performance of modern convolutional neural networks (CNN) and a linear support vector machine (SVM) with respect to spatial contrast sensitivity. Specifically, we compare CNN sensitivity to that of a Bayesian ideal observer (IO) with the signal-known-exactly and noise known statistically. A ResNet-18 reaches optimal performance for harmonic patterns, as well as several classes of real world signals including faces. For these stimuli the CNN substantially outperforms the SVM. We further analyzed the case in which the signal might appear in one of multiple locations and found that CNN spatial sensitivity continues to match the IO. However, the CNN sensitivity was far below optimal at detecting certain complex texture patterns. These measurements show that CNNs can have very large performance differences when detecting the presence of spatial patterns. These differences may have a significant impact on the performance of an imaging system designed to detect low contrast spatial patterns.

# Introduction

Deep convolutional neural networks - comprising a stack of computational layers connected by simple non-linearities - have become an important computational tool. The network parameters are established by training. Much of the excitement in the field arises because the generalization can capture semantic categories, such as the texture of leather or a human face, with an accuracy that far exceeds prior art and matches human accuracy on noise-free, undistorted images (Dodge and Karam 2017; Geirhos et al. 2017). Furthermore, region proposal networks can locate the position of objects within these semantic categories anywhere in an image (Krizhevsky, Sutskever, and Hinton 2012; He et al. 2015b; Ren et al. 2015b). Convolutional and related network learning methods are also being applied to other image systems objectives, including denoising, image reconstruction, super-resolution, object detection, part inspection and camera co-design (McCann, Jin, and Unser 2017; Ledig et al. 2016; Karras et al. 2017; Jain and Seung 2009; Jackson et al. 2017; Schlemper et al. 2017; Liu et al. 2019).

When using a tool to analyze or design an imaging system, it is important to understand the limitations of that tool. There are well-established methods for defining the spatial sensitivity of many critical components of imaging systems, such as lenses, pixel geometry, photon noise and electrical noise (Holst 1998). We introduce a new method that assesses the spatial sensitivity limits of the CNN component of an imaging system. We propose comparing the performance of the CNN to the performance of an ideal observer (equivalently, the Likelihood Ratio test described by the Neyman-Pearson Lemma). The ideal observer (IO) has a rigorous formal definition for the signal-known-exactly and noise-known-statistically case. We implement system tests by creating stimuli with known signals and noise and we then compare the CNN performance with the IO. We also compare the performance of another important but simpler machine learning algorithm, the support vector machine (SVM).

We find that a CNN has relatively higher sensitivity to certain types of spatial patterns than others. The CNNs we assess reach ideal sensitivity for some patterns, but for other patterns sensitivity is 5x lower and even lower than the SVM sensitivity. The approach we introduce and the experiments we describe should be helpful in assessing the CNN component of an imaging system for detection applications in vision science, astronomy, and medical imaging (Wandell 1995; Starck and Murtagh 2001; Zhou, Li, and Anastasio 2019).

# Contributions

- We show that a modern CNN (ResNet) can be trained to detect certain spatial stimuli in the presence of Poisson noise (harmonics, faces, others) at an accuracy level that matches the sensitivity of an ideal observer.

- For other stimuli (certain textures) the asymptotic CNN performance remains substantially lower than the ideal observer or even SVM performance; performance is best for stimuli with high spatial correlation.
- When the stimulus may appear at one of multiple discrete locations, CNN performance can also match the ideal observer.
- We show that the detection performance differs between CNNs, and the spatial sensitivity of a CNN can meaningfully impact the performance of an imaging system.

# Methods

## Image simulation

Test and training images were created using a simulation of a simple camera with diffraction-limited optics and a sensor with Poisson noise. The sensor images were calculated using the open-source and freely available software, ISETCam[1] (Farrell et al. 2003; Farrell, Catrysse, and Wandell 2012; Farrell and Wandell 2015). Unless stated otherwise, the stimuli were simulated as being presented on a uniform background with a mean level of about 300 photons per pixel per capture. This level is typical of many imaging applications.

CNN sensitivity was analyzed with an input-referred measurement: We calculated stimulus detection accuracy for a range of logarithmically spaced performance levels, sweeping out a performance versus contrast curve. We then estimate the contrast level needed to obtain 75% correct detection in an present-absent discrimination. For most spatial patterns contrast was defined as the peak stimulus intensity minus minimum intensity divided by twice the mean intensity. In some cases, the contrast was defined by the standard deviation of the spatial pattern.

### Harmonics and textures

The inputs to the CNN were simulated image sensor data. The simulations calculated a camera's sensor response from a planar scene defined by its spatial-spectral radiance (e.g., a harmonic pattern at some contrast, frequency, phase and orientation). The scene has a horizontal field of view of 10 deg, sampled at 512 rows and columns, and 31 wavelengths (400-700 nm with 10 nm spacing). We modeled the imaging lens as diffraction limited (f/# = 4) with a focal distance of 3.9 mm. The monochrome sensor was ideal (no electronic noise) with a pixel size of 2.8 microns, approximately equal to the full-width half maximum of the diffraction limited lens (2.4 microns). In this configuration the 10 deg scene spans 238 x 238 sensor pixels and the Nyquist sampling frequency for the sensor is approximately 119 cycles/image. The sensor image data include only Poisson noise, which is the classic description of photon absorptions in an electronic device (Schottky 1918).

---

[1] https://github.com/iset/isetcam

## Face stimuli

Face images were taken from the MIT-CBCL database[2] (Weyrauch et al. 2004). We converted these images to a contrast image (mean of zero) and added each to a uniform gray background. The face contrast was measured by its standard deviation, and set to 0.7071, which matches the mean and standard deviation of a harmonic pattern with a contrast of one. We simulated presenting this monochrome image on a display monitor in which each pixel emits an equal photon spectral radiance. The scene radiance was adjusted so that the mean number of photons captured by each pixel was close to 300.

## Cellular automaton textures

We generated complex textures using a cellular automaton method (Wolfram 1983). We scale the scene resolution to 256 x 256, the resolution of the automaton we create, and slightly increase the lens field of view. This way, each pixel within the scene reaches exactly one pixel of the simulated sensor. For the textures the mean and standard deviation of the images were adjusted as we did for the face stimuli (scene radiance standard deviation of 0.7071; mean scene radiance set to create an average of 300 photons per pixel).

## Ideal observer

The neural network was compared to an ideal observer with signal-known-exactly and background-known-statistically. The number of electrons at each position is given by a Poisson distribution (Snyder and Miller 1975), whose rate parameter $\lambda$ is equal to the intensity of the signal at each position in the image:

$$P(N) = \frac{\exp(-\lambda)\lambda^N}{N!}.$$

The ideal observer chooses the more likely signal based on a maximum likelihood calculation. For a candidate signal in noise, $\theta$, measured independently at each pixel, the likelihood is the product of the Poisson density scaled by the a priori likelihood of the signal:

$$L(\theta) = P(\theta) \prod_{i=1}^{p} (P(N_i|\theta).$$

For computational simplicity it is usual to calculate the log likelihood:

---

[2] http://cbcl.mit.edu/software-datasets/heisele/facerecognition-database.html

$$LL(\theta) = \log(P(\theta)) + \sum_{i=1}^{p} \log(P(N_i|\theta)).$$

(Equation 1)

No training is necessary to implement the ideal observer. When there are N different signals, the system selects the most likely of these given the data. This algorithm performs optimally given the available information (Wilson S. Geisler 2011).

## Support Vector Machine

Support vector machines (SVMs) were introduced by (Corinna Cortes and Vapnik 1995) under the slightly different name 'Support-Vector Networks'. The widely used linear SVM uses training data to learn a support vector such that the value of the inner product between this vector and a data sample decides the classification (e.g., signal vs. noise). A linear SVM separating two classes implicitly defines a hyperplane separating the two classes. To solve nonlinear classification tasks it is possible to use a nonlinear kernel, which is an extension of the dot product, as described by (Aizerman, Braverman, and Rozonoer 1964).

We use the linear support vector classifier implementation by the Python library Scikit-learn (Pedregosa et al. 2011), based on the libsvm implementation (Chang and Lin 2011). The SVM classifier optimizes for the hinge loss (Rosasco et al. 2004), which finds the maximum margin classification. The SVM is optimized via a SMO-type decomposition method proposed in (Fan, Chen, and Lin 2005). We set the maximum iterations performed to 1000, unless the convergence tolerance criterion of 0.001 is reached (Chang and Lin 2011).

## Convolutional neural network

We used a ResNet network architecture because of its high quality (He et al. 2015b). The ResNet comprises multiple modules that each perform a convolution, batch normalization, and nonlinear operation (rectified linear unit). The network also includes skip connections.

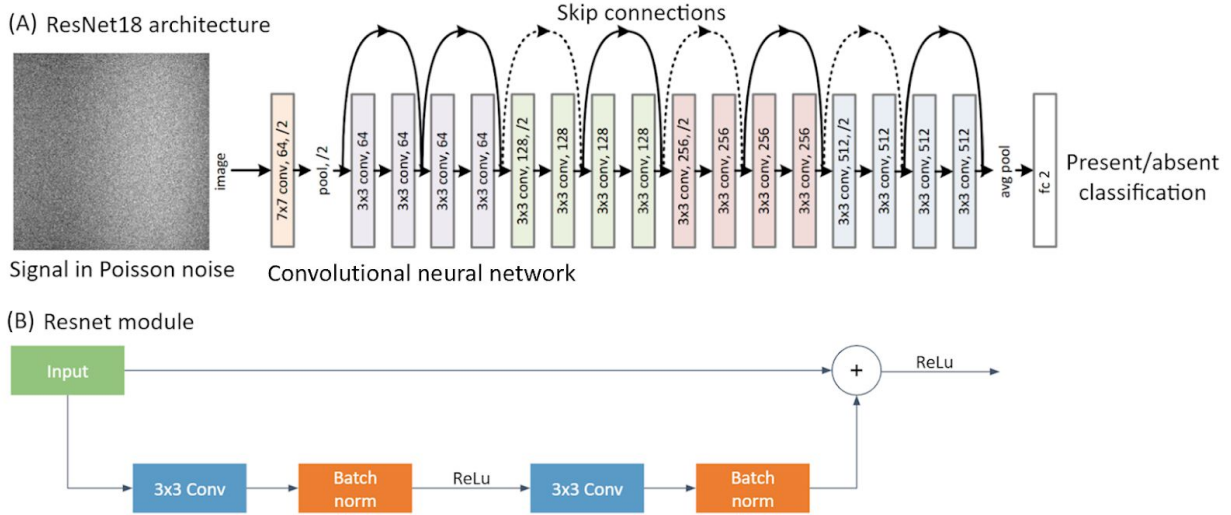

**Figure 1**. The ResNet CNN architecture (He et al. 2015b). (A) The input is processed through 18 stages; many of these stages include connections that transmit the input to a later stage through a skip connection (dashed implies resizing). The final stage is a fully connected layer that provides a classification decision – signal present or absent. The text in each of the stages describes its key properties: N x N conv is the kernel size; the next integer is the number of kernels; if present /N describes the spatial sub-sampling (stride). In our implementation the last fully connected layer only has 2 output classes (signal vs. noise) though in many applications this layer can be quite large. (B) The key concept of the network can also be described as comprising 8 modules. Each module performs a standard set of operations: convolution, batch normalization, half wave rectification (ReLu), convolution, batch norm, skip connection sum, and ReLu.

If not declared differently, the architecture was a ResNet-18 (He et al. 2015b), which has a good trade-off between speed and accuracy for our experiments. We use the PyTorch (Paszke et al. 2017) implementation with a few minor adjustments. We changed the first convolution layer to account for the fact that the sensor data are monochrome. We also replaced the average pooling layer through the PyTorch implementation of an adaptive average pooling layer (Lin, Chen, and Yan 2013) to allow the network to be more flexible to variations in image size. The network weights were randomly initialized by the default PyTorch initialization method, a method known as He Initialization (He et al. 2015a). This algorithm specifically addresses rectifier nonlinearities. The last layer of the ResNet-18, a fully connected layer, is replaced by a smaller layer to accommodate the very small output dimension (binary choice).

The data consists of one scene per class that has random Poisson noise. There is no inherent limit to the epoch size and this parameter can be set arbitrarily. We used 10,000 samples to define one epoch. The batch size was 32, and we used Adam (Kingma and Ba 2014) as the gradient-based optimization function.

The outputs of the neural network are normalized into a probability distribution via the softmax function. These processed outputs are then used to calculate the loss function. For this, we use cross-entropy loss where $y_c$ is the ground truth and $\hat{y}_c$ is the model output for class c.

$$L(y, \hat{y}) = -\sum_{c=1}^{M} y_c \log(\hat{y}_c)$$

The initial learning rate is 1e-3 and after 10 epochs, the learning rate is decreased to 1e-4. After another 10 epochs, the CNN is trained with a learning rate of 1e-5. The network's performance is tested on 5,000 data samples. Seeds are used to ensure the same random initialization of ResNet-18 on all experiments. Training data are generated with the same, specified, seeds to initiate the random number generator.

The ResNet-18 is trained using a parallel algorithm to permit the server to use all available GPUs. While each neural network runs on one specific GPU, each GPU runs multiple training and testing experiments in parallel. On the server used for training, there are six Nvidia GK210 graphics processors. On one GPU, training ResNet-18 with 300,000 data samples, generated in real-time, takes 1:18 hours.

In the Appendix we briefly present the results using two other networks: VGG-16 and AlexNet. For very simple spatial patterns all networks perform at similar levels. For more complex signals the networks differ substantially (Figures A2 and A3).

## Network performance

### Metrics

The detection experiments are two-class classification problems. We vary the size of the signal contrast, position, or orientation and measure classification performance by the hit (true positive) false alarm rates of the IO, ResNet-18, and SVM. The network training was carried out for each stimulus at each contrast level. We estimate the discriminability between the two classes using d-prime ($d'$) (Stanislaw and Todorov 1999; Green and Swets 1988) from these two rates. Specifically, we calculate the z-scores (inverse of the standard normal cumulative distribution) for these rates and subtract the false alarm z-score from hit rate z-score:

$$d' = Z(hit\,rate) - Z(false\,alarm\,rate).$$

We manage extreme hit or false alarm rates (zero errors) by a small adjustment to the hit and false alarm rates (Knoke, Burke, and Burke 1980):

$$hit\,rate\ =\ \frac{0.5+\sum hits}{1+\sum hits+\sum misses},$$

$$false\,alarm\,rate\ =\ \frac{0.5+\sum false\,alarms}{1+\sum false\,alarms+\sum correct\,rejections}.$$

Without this modification, a hit rate of 100% would result in a $d'$ of infinity, given the false alarm rate is not at 100% as well. The modified equations for false alarm and hit rates provides a finite and only slightly biased underestimate of the true $d'$ (Hautus 1995).

Given the mean number of signal photons in a detection task with only Poisson noise, $d'$ can also be calculated by a formula that only requires the mean photon absorptions of both classes. In this formula, the sum of scaled Poisson random variables is approximated with normal density (W. S. Geisler 1984):

$$d' = \frac{\sum_{i=1}^{n}(\beta_i - \alpha_i)\ln(\beta_i/a_i)}{[0.5\sum_{i=1}^{n}(\alpha_i + \beta_i)\ln^2(\beta_i/\alpha_i)]^{1/2}}.$$

Our results show that the IO $d'$, calculated via hit and false alarm rate, matches the theoretical $d'$. We also calculate the sensitivity of a discriminator.

In most analyses, we calculate how $d'$ increases as the stimulus parameters - contrast, position shift, or angle - change. This produces a curve relating performance ($d'$) to the stimulus parameter. In many analyses we summarize network sensitivity using an input-referred measure. Specifically, we calculate the contrast level, spatial shift or orientation angle needed to achieve $d'$ = 1.5. The contrast, phase shift or angle metric is calculated by linearly interpolating the performance curve.

All experiments were performed in a seeded, deterministic manner; we also examine the influence of random components by carrying out certain experiments multiple times, varying the seeds. This results in different training and test data, different neural network random states, as well as varied random states of the SVM. In the graphs below with such replications we repeated the experiments with different seeds five times, and we report the mean and standard deviation of the sensitivity measure.

### Size of training data

The ResNet-18 performance improves as training set size increases (Figure A1). The ResNet-18 reaches asymptote - in this case the maximum theoretical performance level - when the training set reaches 100,000 to 300,000 samples. The SVM performance reaches asymptote at a much smaller training set size, prior to the initial portion of the graph (about 10,000 training samples).

The ResNet-18 performance is significantly better than that of the SVM after 10,000 training samples, continuing to rise up to the IO level at approximately 100,000 training samples. Based on these experiments, we used a training set size of 10,000 samples for the SVM and 300,000 samples for ResNet-18.

# Results

First, we consider the detection of harmonics. We measure discriminability as a function of spatial frequency, position (phase shift) and orientation. Second, we measure signal detection based on signal size (disks of various sizes). Third, we consider a collection of biological images (faces). Fourth, we measure texture signals that are not compact in space or spatial frequency (white noise, cellular automata). Fifth, we analyze the detection performance for targets in which the signal may be present in one of multiple positions.

## Harmonics

### Contrast

For all networks detection sensitivity ($d'$) of a harmonic in Poisson noise increases with contrast. The ResNet-18 can be trained to achieve a performance that closely matches the ideal observer's performance and the SVM performance is about half a log (4x) unit less sensitive (Figure 2A).

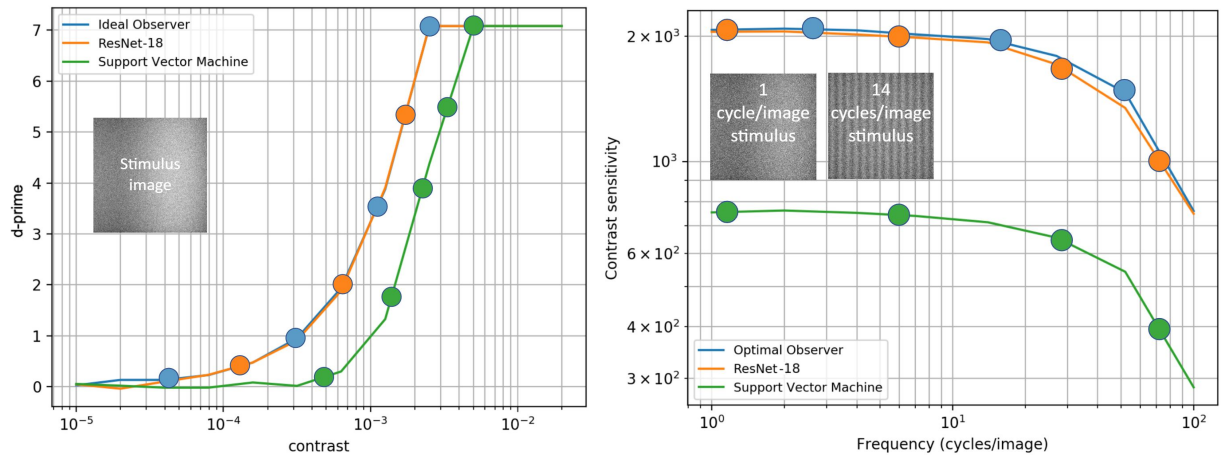

**Figure 2.** (A) Comparison of detection performance ($d'$) for a harmonic presented in Poisson noise. Performance increases as a function of contrast. (B) Contrast sensitivity functions of the IO, ResNet-18 and SVM for spatial frequencies up to the sensor Nyquist frequency. The sensitivity is defined as the inverse of the contrast needed to achieve discrimination performance of $d'$ = 1.5. Higher contrast sensitivity means performance is reached with less contrast.

We repeated these calculations for a range of harmonic spatial frequencies, extending to the Nyquist limit of the sensor (Figure 2B). The input-referred contrast sensitivity (1 over the contrast for $d'$=1.5 ) matched the performance of the ideal observer closely, being only only slightly lower than the IO, by an average of 2.86% (0.013 log10 units). The SVM contrast sensitivity was an average of 63.39% lower (0.44 log10 units) compared to IO.

## Disks

Detection sensitivity grows systematically with disk radius, approximately as the square root of the disk area (Figure 3). Deviations from this rule are present for small disks which are blurred by the optics and very large disks that span nearly the whole sensor. The ResNet-18 again approximates IO performance for all disk sizes tested, and the SVM is about half a log unit lower.

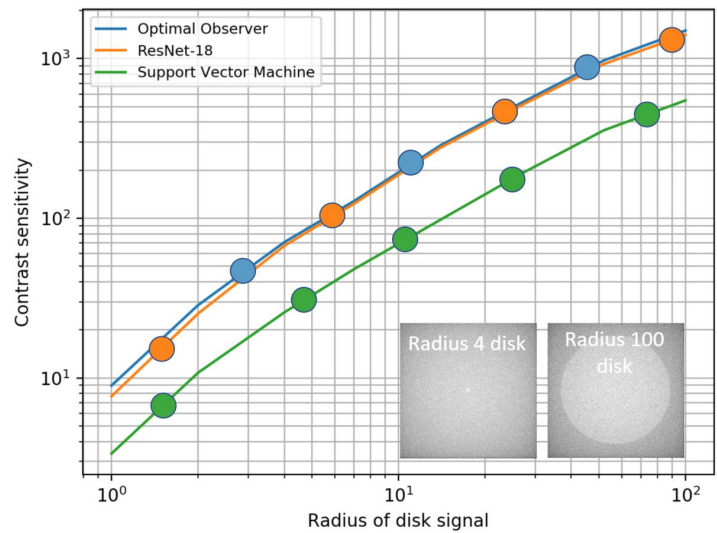

**Figure 3.** Contrast sensitivity to disks for IO, ResNet-18 and SVM. Detection performance for disks with sizes from radius 1 to radius 100. Disk contrast sensitivity is shown for a performance level of $d'$ = 1.5.

## Faces

Disks and harmonic signals are very simple patterns compared to many natural objects. Hence, we decided to measure contrast sensitivity for an important and complex object, the human face (Figure 4).

The ResNet-18 contrast sensitivity is similar but slightly lower than the IO sensitivity. ResNet-18 contrast sensitivity is on average 5.87% lower than the IO sensitivity. This difference is slightly larger than the sensitivity difference using the harmonics. The SVM performance is about 1/3rd the sensitivity of the IO and ResNet-18 network.

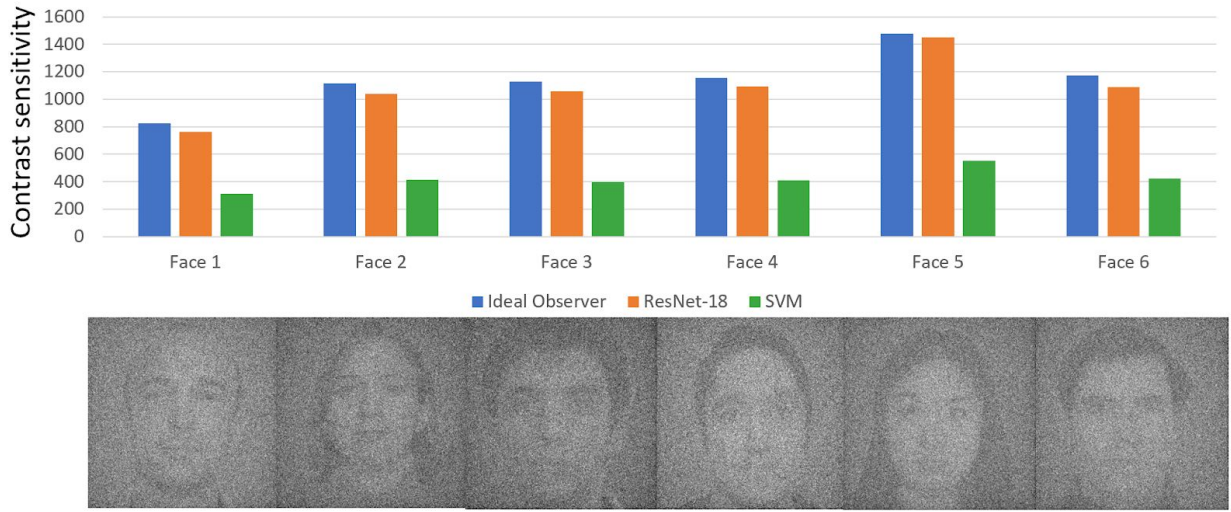

**Figure 4.** Contrast sensitivity to faces for IO, ResNet-18, and SVM. The first five graphs show detection of a single face. The sixth graph shows sensitivity to a collage comprising nine faces. Contrast sensitivity is shown for a performance level of $d'$ = 1.5.

# Textures

In addition to test stimuli (harmonics, disks) and natural objects (faces), there are applications in which the target is a texture pattern (see Discussion). We used cellular automata to generate an organized list of texture patterns (Wolfram 1983, 2016). We focused on rules which converge to a structured repetitive pattern (class 2) and rules in which the texture pattern remains random (class 3). We generated textures using four different class 2 rules and four different class 3 rules.

## Class 2 cellular automata

Class 2 automata converge to a repetitive texture pattern. We suspect that CNNs might learn filters to identify repetitive patterns. To measure detection performance, we used experiments for four class 2 automata (rules 3, 57, 76 and 78). The contrast sensitivity for these patterns is slightly higher for the IO than ResNet-18, and substantially higher than SVM (Figure 5A).

Slightly worse performance is achieved for the other two automata. At rule 3, IO has a contrast sensitivity of 1213.31, while ResNet-18 reaches 861.14 and SVM achieves 438.02. IO contrast sensitivity for the rule 57 automaton is the lowest. Here, IO reaches 824.34, ResNet-18 achieves 688.76 and SVM reaches 298.32. Compared to IO, ResNet-18 performance drops by an average of 18.18%, while SVM performance drops by an average of 63.74%.

## Class 3 cellular automata

Class 3 automata have a complex irregular pattern (Figure 5B). We examine four class 3 automata (rules 22, 30, 75 and 101). The CNN sensitivity to these patterns is far from the sensitivity of the IO, dropping to the level of the SVM performance.

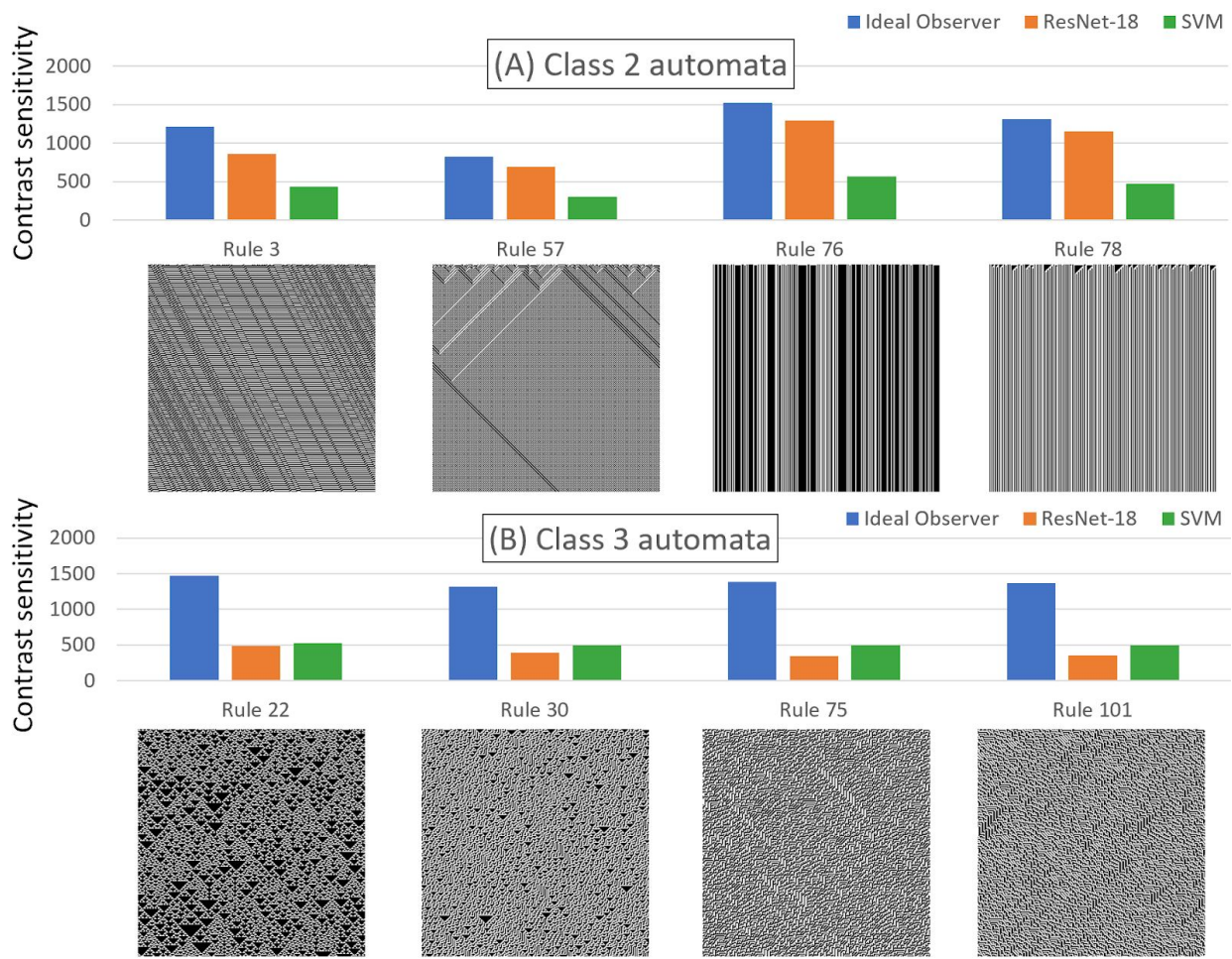

**Figure 5.** Contrast sensitivity for class 2 cellular automata (A). ResNet-18 contrast sensitivity is lower than IO; the sensitivity of SVM is around one third of IO sensitivity. The highest sensitivity is for rule 76, followed by rule 78. One-dimensional patterns are easiest to detect. Contrast sensitivity for class 3 cellular automata (B). IO contrast sensitivity is three-fold higher than either ResNet-18 or SVM. In several cases, SVM contrast sensitivity to these patterns exceeds that of ResNet-18. Unlike the class 2 cellular automata, these textures are dense and not space-invariant. The variance of the contrast sensitivity is smaller than the measured difference in contrast sensitivity.

## Block randomization

In addition to the cellular automata, we produced texture patterns by randomizing the pixel positions in an existing image. We performed a series of experiments by block-wise scrambling the pixels in a one-cycle per image harmonic (Figure 6).

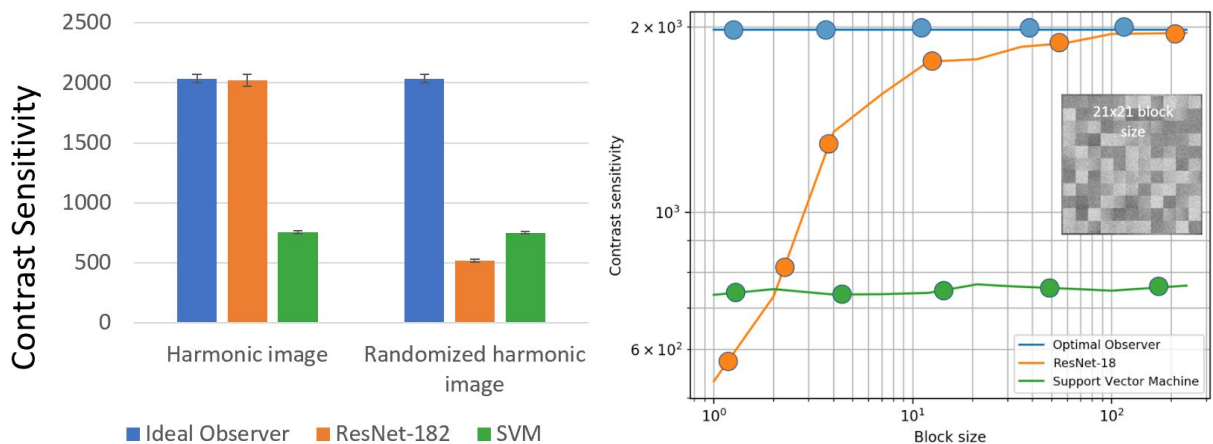

**Figure 6.** Performance for spatial randomization of frequency one harmonic signal. Panel (a) shows 70% reduction in ResNet-18 contrast sensitivity (compared to IO) for randomization of all pixel locations of harmonic signal (1x1 block). The contrast sensitivity is 20% lower than the SVM. The bar heights represent the mean of five runs with training data, test data and different random number seeds. Error bars are +/- 1 SD. Panel (b) displays contrast sensitivity for various block sizes.

The IO performance is indifferent to the scrambling, as expected from the computational formula (Equation 1). Similarly, the SVM adjusts its critical vector and learns to detect the pattern with reordered pixels. The ResNet-18 sensitivity is substantially reduced by scrambling. The scrambled texture pattern does not repeat regularly across the image, and like the cellular automata in class 3, the ResNet-18 sensitivity is below the IO.

## Multiple target positions

The ability to detect and localize a signal anywhere in a scene is one of the most important contributions of CNN technology (Ren et al. 2015a). We compare the CNN sensitivity with the ideal observer sensitivity to a simple stimulus (a Gabor patch) that might be presented at one of multiple possible locations (Figure 7). When there are N different locations, the ideal observer selects the most likely of these locations, or no signal, given the image data.

Introducing position uncertainty reduces the sensitivity of the ResNet, SVM and the IO. Although sensitivity declines, the ResNet-18 continues to match the IO performance. Both methods are about half as sensitive when the target can appear in 16 locations rather than one location. The SVM sensitivity declines by a larger fraction, becoming about one-fourth as sensitive as the number of possible positions increases to 16 from one.

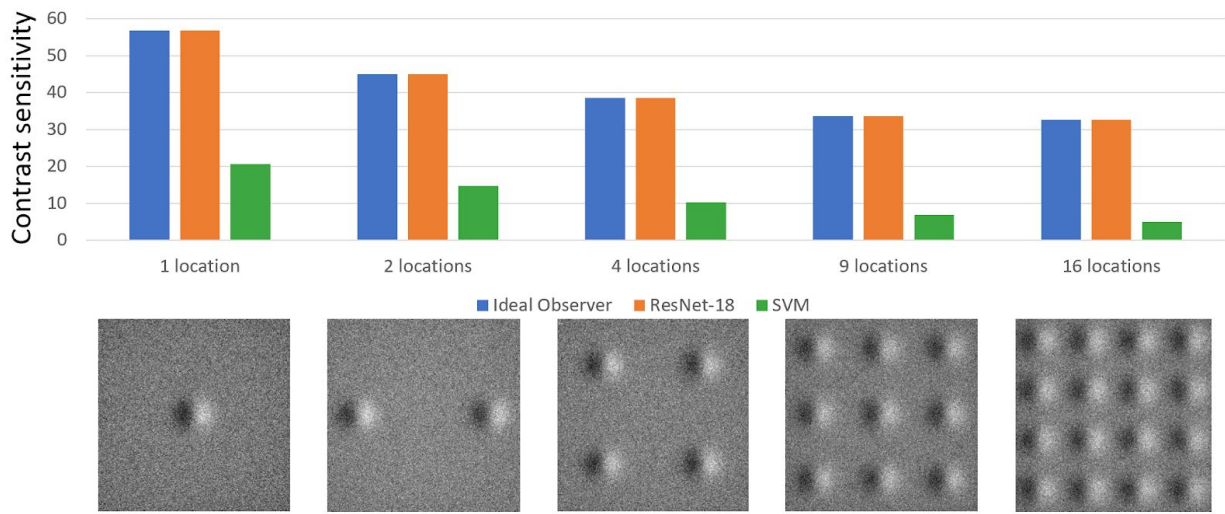

**Figure 7:** Detection performance of frequency one harmonic with Gabor for one or multiple locations. The signal examples show the signal in all its potential locations. In the signal case, the signal can be seen in exactly one location. The bar heights represent the mean of five runs with training data, test data and different random number seeds. Error bars are +/- 1 SD.

## Comparison to VGG-16 and AlexNet

We compared the performance of ResNet-18 to two other CNNs, VGG-16 (Simonyan and Zisserman 2014) and AlexNet (Krizhevsky, Sutskever, and Hinton 2012). We measured contrast sensitivity using a harmonic with one cycle per image ( Figure A2, cf. Figure 2). The VGG-16 and AlexNet sensitivities are close to the IO for higher contrasts. But in both cases the IO performs above chance ($d'$ > 0) at contrasts where the two networks are still at $d'$ = 0. Even at the higher contrasts AlexNet sensitivity is slightly lower than IO sensitivity.

Next we explored network sensitivity to variations of the harmonic signal, the multiple faces signal, and two different cellular automata (Figure 8). All network hyperparameters were the same as for ResNet-18 with one exception: for the network solution to converge it was necessary to decrease the VGG-16 and AlexNet initial learning rates to 1e-5 rather than using the ResNet initial learning rate of 1e-3. It might be possible to find further improvements in VGG-16 and AlexNet performance by modifying other hyperparameters.

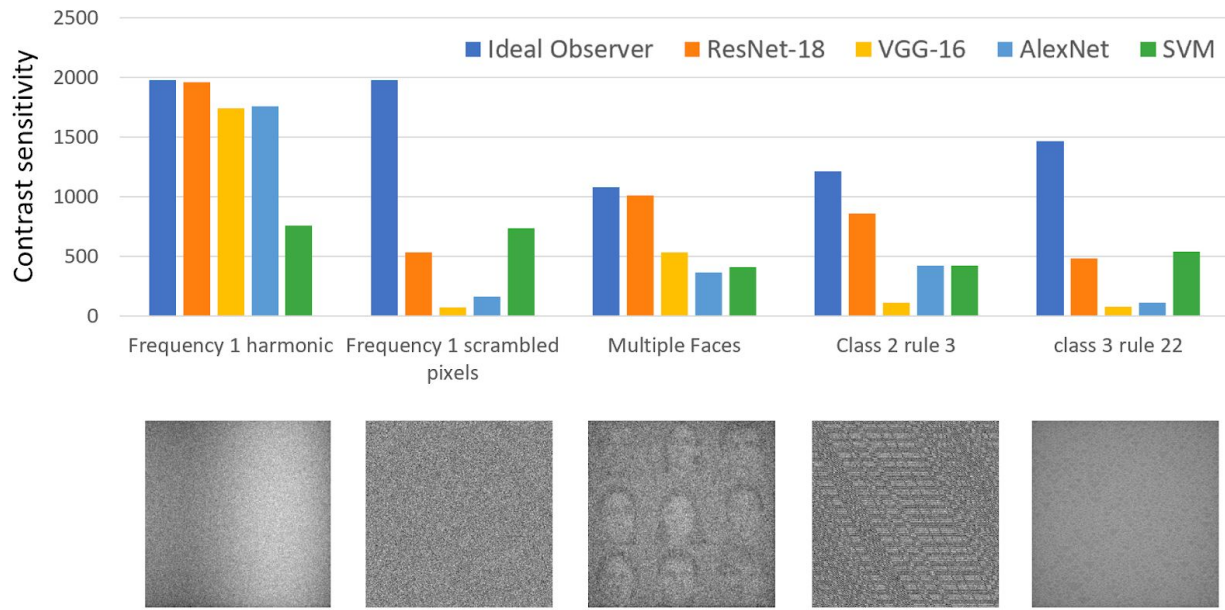

**Figure 8:** Detection sensitivity of IO, ResNet-18, VGG-16, AlexNet and SVM to six different stimuli that were also used in the main text. See the text for details.

Like the ResNet, randomizing image pixel positions (1x1 blocks) causes VGG-16 and AlexNet sensitivity to drop significantly. The sensitivity of these two networks is also substantially lower for detecting multiple faces signal and the two types of automata signals. In several cases the SVM outperforms the VGG-16 and AlexNet CNNs.

# Discussion

## CNN spatial sensitivity

For many spatial patterns (harmonics, disks, faces), a ResNet-18 can be trained to detect a contrast pattern at the same sensitivity level as an ideal observer. The main requirement to achieve this optimal performance is a large number of training samples (more than 1e+5 samples).

The ResNet-18 spatial sensitivity for certain textures (class 3 cellular automata and block-scrambled images) is 2.5x lower than the ideal observer and comparable to the SVM. All of the networks have reduced sensitivity to these patterns. Network sensitivity to the repetitive textures, such as class 2 automata is higher than sensitivity to random class 3 automata textures.

There is also a substantial spatial sensitivity difference for the two other architectures (VGG-16 and AlexNet). For these architectures both class 2 and class 3 textures are detected poorly compared to ideal, and sensitivity to faces is only half that of the IO (Figure 8). It is worth noting that individual networks have their distinct spatial sensitivity profiles.

## Uncertain signal position

An important value of CNNs is their ability to detect patterns even when the pattern's position is uncertain. We performed an initial analysis of the ResNet's ability to detect signals present at one of multiple positions and found that the CNN matches IO contrast sensitivity. The experiments examining the sensitivity when position is uncertain could be significantly expanded to include variations in the stimulus pattern, size and a systematic analysis of position bias. The methods in this paper - input-referred contrast measures and a comparison with the ideal observer - can provide meaningful numerical assessments for such evaluations.

## Architecture

We compared ResNet-18 detection sensitivity to other well-known CNN architectures, VGG-16 and AlexNet. Sensitivity to harmonics is comparable, but sensitivity to more complex signals (e.g., faces) is substantially lower for AlexNet and VGG-16; in several cases these networks are less sensitive than a linear SVM.

ResNet-18 contrast sensitivity is lower than the IO sensitivity when the stimuli comprise fine textures that do not repeat regularly across the image. Block randomization and cellular automata examples fit this pattern. Why limits performance on the class 3 cellular automata and block-scrambled images? The central difference between cellular automata of class 2 vs. class 3 and block-scrambled images is the image complexity. The repeating patterns of class 2 automata can be summarized by a shift-invariant representation compared to the non-repetitive class 3 and block-scrambled patterns. The stimulus structure's complexity matches the architecture of the CNN, which has a small number of weights compared to fully connected NNs. The number of weights needed by the IO to distinguish the signal from noise using is (256 x 256 x 2, row x col x classes). The initial stage of the ResNet-18 uses only 7x7x64 weights for low-level feature extraction which is just 2% of the IO weights. The total number of ResNet-18 weights is vastly larger (1.1e+7).

## Applications

There are signal detection applications whose signals resemble class 3 cellular automata (MRI k-space, which is the Fourier Transform of the image, skin rashes or retinal bleeding). The conventional CNN architectures limit performance for such signals. The tests in this paper can discriminate between different CNN architectures to determine which may be most effective for specific classes of signals. An advantage of our testing procedure is that it does not require large amounts of labeled data which is especially helpful for signal types that are only observed in certain clinical cases.

In addition to the benchmarking of existing CNN architectures, we hope that our tools will furthermore be helpful in the design of new, innovative CNNs that allow improved performance on non-standard signal types.

## Conclusion

We present a way to assess CNN performance by measuring performance with respect to a fundamental image science tool, the ideal observer. This allows us to quantify how well a CNN architecture can learn to detect signals of varying shapes and abstraction. As in other branches of image systems engineering, we hope that analyzing CNN architectures will help CNN designers to find the right modifications that optimize deep learning algorithms for specific tasks. Just as we characterize the impact on spatial resolution of the lens, pixel sampling array, and electrical noise, we should characterize the impact of a CNN detection network.

ResNet, along with many other CNN architectures, was designed for image classification and commonly tested with the categories in ImageNet. Compared to fully connected neural networks, the CNN architecture processes low-level features via convolutional filters which reduces the number of weights required. We evaluated several CNNs for signal detection sensitivity. For many signals we find that even in the presence of pixel-wise Poisson noise the ResNet CNN has the same sensitivity as an ideal observer. We conclude that current CNN architectures are able to detect signal types, similar to ImageNet motifs, at near optimal levels.

Image systems can be designed to detect a wide range of spatial targets in applications spanning medical imaging and industrial inspection: from localized tumors and moles to a widespread rash. Not all of these targets are similar to the image types used by ImageNet, the task on which most of these CNN architectures are benchmarked. We hope that our signal detection analyses will add to the understanding on how well these architectures perform on signals that differ from this standard.

The high sensitivity of a CNN for identifying certain targets, but not others, should be a part of decision-making in image system design.  The experiments in this paper are a start towards developing this technology.  It would be useful to develop consensus methods that assess the spatial sensitivity profile of a CNN with respect to the target objects for each application.

# Appendix

## A1. Accuracy based on size of training set

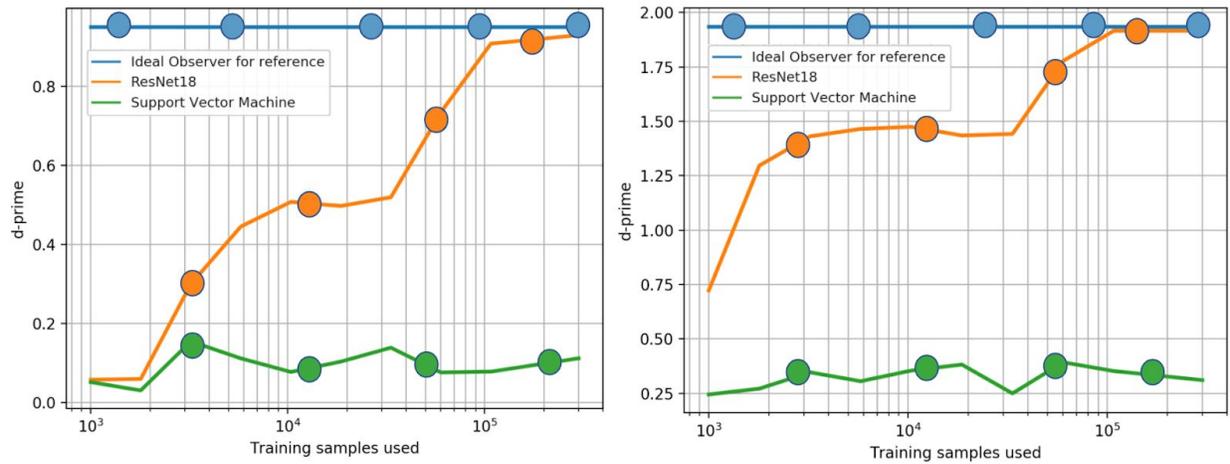

**Figure A1.** Increase in detectability ($d'$) of a harmonic image in Poisson noise as a function of the number of training samples (horizontal axis). The two panels show performance for stimuli at two different contrasts: 3.2e-4 (left) and 6.3e-4 (right). Irrespective of training set size, ResNet-18 was trained for 9375 iterations with a batch size of 32. The ideal observer requires no training. The SVM reaches asymptotic performance before 1e+4 training samples. The ResNet performance increases until approximately 3e+5 training samples. As in the main text, colored disks are superimposed on the lines at every other measurement point, which is helpful when the ResNet and IO curves superimpose.

## A2. Comparison of network contrast-dependent accuracy

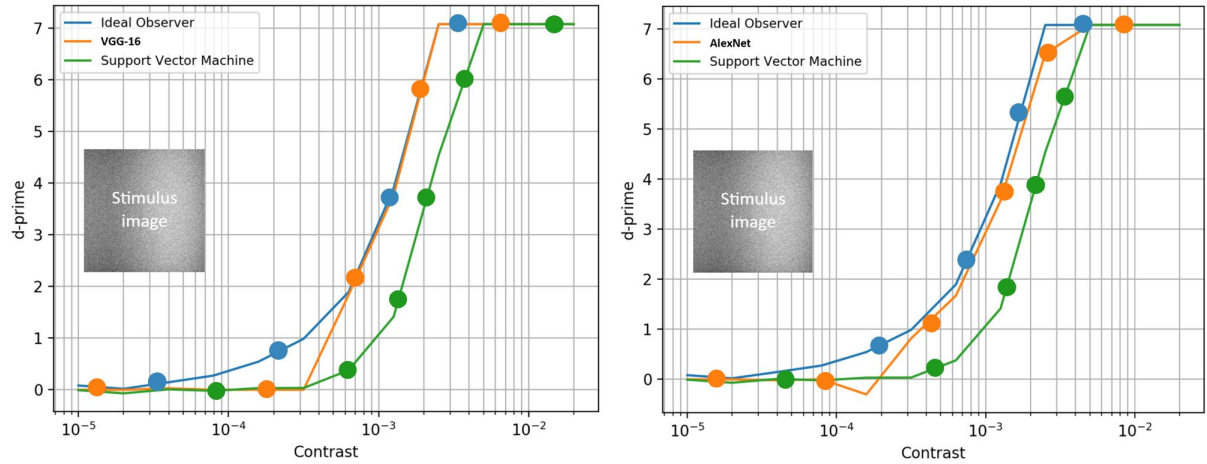

**Figure A2:** Detection performance of VGG-16 (left) and AlexNet (right) for a harmonic stimulus of frequency one. VGG-16 (A) approximates the IO for higher contrasts but does not reach IO performance at lower contrasts. AlexNet (B) is able to discriminate a low contrast signal slightly better than VGG-16, but never reaches IO performance for high contrast harmonic curves.